\documentclass[letterpaper, 10 pt, journal, twoside]{IEEEtran} 
\usepackage{amsmath,amsfonts}
\usepackage{algorithmic}
\usepackage{algorithm}
\usepackage{array}
\usepackage[caption=false,font=normalsize,labelfont=sf,textfont=sf]{subfig}
\usepackage{textcomp}
\usepackage{stfloats}
\usepackage{url}
\usepackage{verbatim}
\usepackage{graphicx}
\usepackage{cite}
\usepackage{multirow}
\begin{document}

\title{Cross-Category Functional Grasp Transfer}

\author{Rina Wu$^{1}$, Tianqiang Zhu$^{1}$*, Xiangbo Lin$^{1}$,Yi Sun$^{1}$*~\IEEEmembership{Member,~IEEE}
\vspace{-0.5cm}

\thanks{Manuscript received 6 June 2024; accepted 23 September 2024. This letter was recommended for publication by Editor Júlia Borràs Sol upon evaluation of the Associate Editor and Reviewers’ comments. This work was supported by the National Natural Science Foundation of China under Grants 62373075 and 61873046. (Corresponding authors: Yi Sun, Tianqiang Zhu)}

\thanks{The authors are with the School of Information and Communication Engineering, Dalian University of Technology,Dalian, 116024, China.
        {\tt\small hswrn@mail.dlut.edu.cn,zhutq@mail.dlut.edu.cn, linxbo
@dlut.edu.cn,lslwf@dlut.edu.cn}}
\thanks{The project website is: https://github.com/wurina-github/CCFGTransfer.}
\thanks{Digital Object Identifier (DOI): see top of this page.}
}

\markboth{IEEE Robotics and Automation Letters. Preprint Version. Accepted September, 2024}{WU \MakeLowercase{\textit{et al.}}: Cross-Category Functional Grasp Transfer}


\maketitle

\begin{abstract}
Generating grasps for a dexterous hand often requires numerous grasping annotations. However, annotating high DoF dexterous hand poses is quite challenging. Especially for functional grasps, requiring the hand to grasp the object in a specific pose to facilitate subsequent manipulations. This prompts us to explore how people achieve manipulations on new objects based on past grasp experiences. We find that when grasping new items, people are adept at discovering and leveraging various similarities between objects, including shape, layout, and grasp type. Considering this, we analyze and collect grasp-related similarity relationships among 51 common tool-like object categories and annotate semantic grasp representation for 1768 objects. These objects are connected through similarities to form a knowledge graph, which helps infer our proposed cross-category functional grasp synthesis. Through extensive experiments, we demonstrate that the grasp-related knowledge indeed contributed to achieving functional grasp  transfer across unknown or entirely new categories of objects.
\end{abstract}


\section{INTRODUCTION}
\label{sec:intro}

\par \IEEEPARstart{W}{ith} the rapid development of deep learning, the performance of the dexterous grasp of robots has greatly improved through deep learning from a large amount of labeled grasps. However, annotating the high DoF grasps of multi-fingered robot hands on a large-scale objects is laborious and time-consuming. This prompts us to delve deeper into how humans can infer the various grasps of new or unseen objects only through the past grasping experience on limited amounts of objects. Such a skill of grasp transfer is a hallmark of machine intelligence.
\par Previous studies in grasp transfer typically warp contact points on the source object to the target object through global and local shape similarities~\cite{vahrenkamp2016part}, coherent latent shape space~\cite{simeonov2022neural} and dense corresponding descriptors~\cite{yang2021learning}. More recently, several studies transfer labeled grasp poses~\cite{yang2022oakink} or hand-object touch code~\cite{wu2023functional} from only one object to other ones within a category by the 3D-to-3D dense shape correspondence. This is based on the fact that objects of similar shape in a category have similar grasps. Despite these recent successes in the transfer of grasping skills within similar objects, there is still no solution to the transfer of grasping skills to different categories of objects with significant differences in shape and function, which is the focus of this study.
\begin{figure*}
    \centering
    \includegraphics[width=0.92\linewidth]{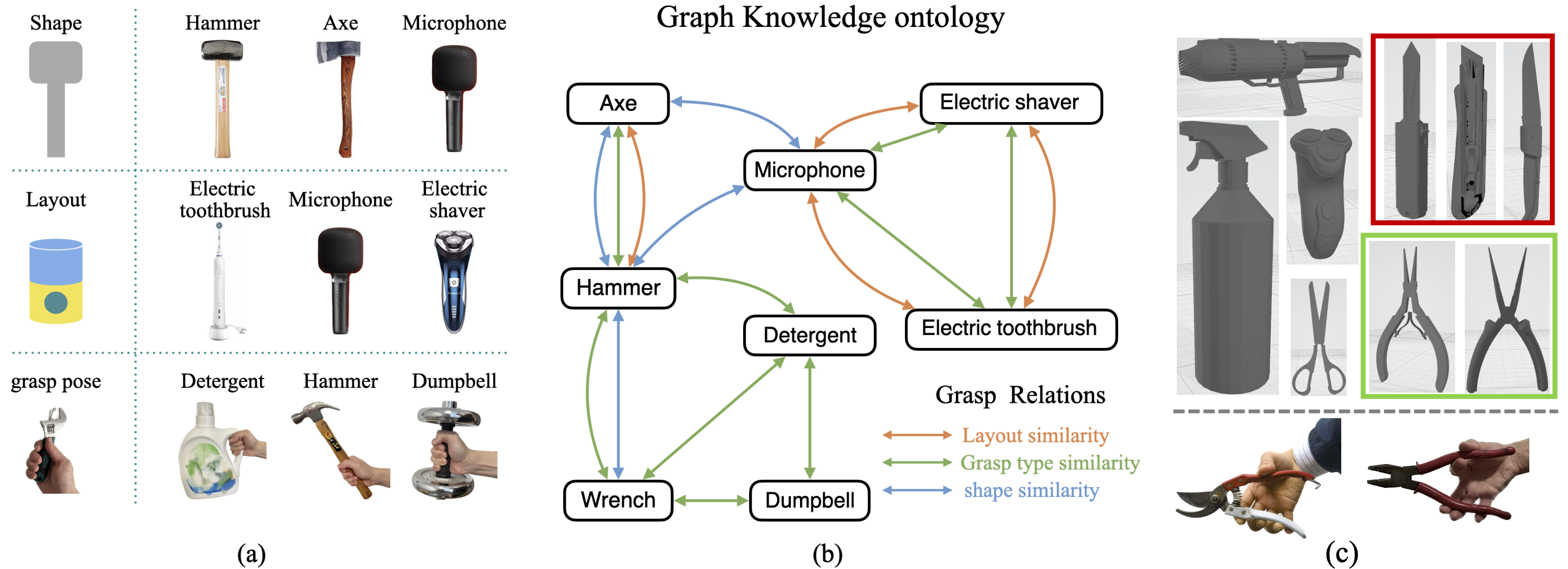}
    \vspace{-0.3cm}
    \caption{(a) Three object attributes related to cross-category grasp transfer, as well as representative objects. (b) The part of knowledge graph ontology constructed based on the similarity of objects' three attributes. (c) Presentation of some objects in the dataset.}
    \vspace{-0.3cm}
    \label{fig_teaser}
\end{figure*}
\par To truly achieve human-level grasp skills for robot, we need to identify how a human grasp can be transferred across multiple categories of objects. In this work, we exploit the grasp-related semantic knowledge of objects to transfer grasps to new classes of objects. We start with 3 main observations. i) \textit{Similar Shape}. Although the shape variance of objects in different categories is greater than that of objects in the same category, there is still shape similarity between some objects, as shown in Fig.~\ref{fig_teaser} between a hammer and a microphone. ii) \textit{Similar Layout}. For some different shapes of objects across categories, their layout of various parts is similar. For example, electric toothbrushes, microphones, and electric shavers have similar layout for functional grasp with the thumb placing on the switch and the remaining four fingers holding the body, as shown in Fig.~\ref{fig_teaser}. iii) \textit{Similar Grasp Pose}. Some objects with significant shape variance may have the same grasp pose, such as a detergent bottle and a hammer in Fig.~\ref{fig_teaser}. To encode these grasp relations between objects, we have established a grasp knowledge graph, as shown in Fig.~\ref{fig_teaser}(b), which can be used in our graph convolutional network-based grasp transfer framework.

\par It should be noted that this letter focuses on more complex functional grasp for post-grasp task manipulation. Hence, instead of directly transferring grasp pose, we transfer semantic touch code~\cite{zhu2023toward}, a type of grasp representation that can be used for synthesizing functional grasps of objects. This is because that the touch code is easier to expand and has stronger generalization than hand-specific grasp pose and contact map~\cite{brahmbhatt2019contactdb}. To better support cross-category functional grasp synthesis and establish new benchmarks for evaluating transfer performance, we increase the object categories of the dataset in~\cite{zhu2023toward} from 18 to 51 and scale the number of objects from 129 to 1768. Additionally, we design a taxonomy for functional grasps to enhance the accuracy of grasp synthesis. By applying grasping relationships to graph convolutional networks, the semantic representation of functional grasp can be transferred to different categories of objects, and functional grasps of these objects can be ultimately optimized under the guidance of their transferred representations. To the best of our knowledge, this is the first effort in demonstrating functional grasp transfer for the multi-fingered hand across different categories of objects. The contributions of this work can be summarized below:
\begin{itemize}
\item{We propose 3 types of grasping relations across different categories of objects, and establish a grasp knowledge graph, which can be used in a graph convolutional network for grasp transfer across categories.} 
\item{We not only scale the grasp dataset~\cite{zhu2023toward} in terms of object categories and numbers, but also add grasp types and grasping relations, which both provide a better benchmark for functional grasp synthesis and cross-category functional grasp transfer.}
\item{We propose a novel cross-category functional grasp synthesis pipeline, which is composed of a touch code transfer framework based on graph convolutional networks and an optimization-based grasp synthesis module. Extensive experiments demonstrate that it can bridge the gap of functional grasp transfer to unseen and even completely new categories of objects.}
\end{itemize}
\vspace{-0.1cm}

\section{RELATED WORK}
\par \textbf{Functional grasp transfer: }Functional grasp is a research direction within task-oriented grasp that focuses on generating appropriate grasp pose by fully considering the subsequent manipulation requirements of the object~\cite{maranci2024enabling}. Existing data-driven methods~\cite{yang2024learning,wu2023functional,yang2022oakink} leverage extensive data to train networks, enabling them to generate functional grasps for previously unseen objects, but building such comprehensive datasets consumes a lot of time and effort. Recently, some studies have started exploring cost-effective grasp transfer methods to enable functional grasp generation for more objects. Some methods~\cite{vahrenkamp2016part,aktacs2022deep,Tekden2023grasp,geng2023gapartnet} divide objects into contactable and non-contactable regions and use correspondences between contactable regions (such as the handles of different hammers) to achieve grasp transfer. However, this coarse division is unsuitable for precise multi-finger grasping in complex tools, like the functional grasping of a drill. To address this,~\cite{rodriguez2018transferring} directly transfers dexterous hand functional grasp poses based on shape similarity within the same category of objects, while~\cite{wu2023functional} uses dense correspondences of surface points between objects of the same category to transfer hand-object touch code~\cite{zhu2023toward}, thereby generating functional grasps for new objects with a dexterous hand. However, the above methods only consider shape similarity within the same object category, which is insufficient for achieving functional grasp transfer across different object categories. In contrast, this paper delves into the mechanism of cross-category functional grasp transfer, proposes three types cross-category similarities, and constructs a network architecture suitable for cross-category grasp transfer.

\par \textbf{Dexterous Manipulation Datasets: }Recently, numerous works have explored multi-fingered hand grasping, focusing primarily on stable grasp based on large-scale synthetic datasets~\cite{liu2019generating,van2020learning,wei2022dvgg,mayer2022ffhnet}. However, for more complex functional grasp, there are only a few datasets available. For example,~\cite{yang2022oakink,liu2022hoi4d} record hand poses during real human-object interactions using visual sensors,~\cite{jian2023affordpose} annotates grasp poses and contact affordance by the simulation software, and~\cite{brahmbhatt2019contactdb} employ thermal camera to record contact map left on object surface after grasping. Moreover,~\cite{zhu2023toward} proposes a semantic grasp representation, touch code, based on the object's own components, which is universal for various dexterous hands. Nevertheless, all the above datasets lack support for cross-category functional grasp transfer. In contrast, our dataset defines associations across various object categories, and includes 51 categories and 1768 objects, covering as many everyday tools as possible, surpassing most existing datasets. 
\section{METHOD}
\label{sec:3}
To tackle the challenge of cross-category functional grasp transfer, we first established grasp-related similarities between objects across categories in Sec.A. In Sec.B, we selected a grasp representation that is suitable for cross-category transfer. In Sec.C, we proposed a method to achieve cross-category transfer of this grasp representation based on object similarity. Finally, we presented how to generate a functional grasp for an object based on the transferred grasp representation in Sec.D.
\vspace{-0.2cm}
\subsection{Grasp Similarities across Different Categories}\label{sec:3.1}  
\par To achieve human-level cross-category functional grasp synthesis, it is essential to comprehend the cross-category associations that support human inference for functional grasping across diverse object categories. To this end, we analyze the factors driving human grasp transfer across multiple categories of objects from three aspects: the global shape of objects, the object layout (i.e. spatial relative position of local parts), and the grasp type. 

\par \textit{Shape similarity.} An object's external shape is its most easily observable attribute. Shape similarity can directly serve as the basis for transferring functional grasps between simple tools, such as from a hammer to an axe. 
\par \textit{Layout similarity.} Although global shape similarity facilitates associations among simple objects, it falls short in capturing connections for objects requiring precise manipulation, like electric shavers and microphones. These objects' cross-category functional grasp inference needs to further consider the local parts, especially the spatial relative positions between various parts (i.e., the layout of the object). For instance, electric shavers, microphones, and electric toothbrushes, despite various shapes, share a consistent functional grasp pose due to the consistent relative position of touchable parts, like the switch on the middle of the handle, as shown in Fig.~\ref{fig_teaser}(a). Therefore, constructing layout similarity across different categories of objects will facilitate the cross-category functional grasp transfer for complex tools.

\begin{figure}[!t]
\centering
\includegraphics[width=1\linewidth]{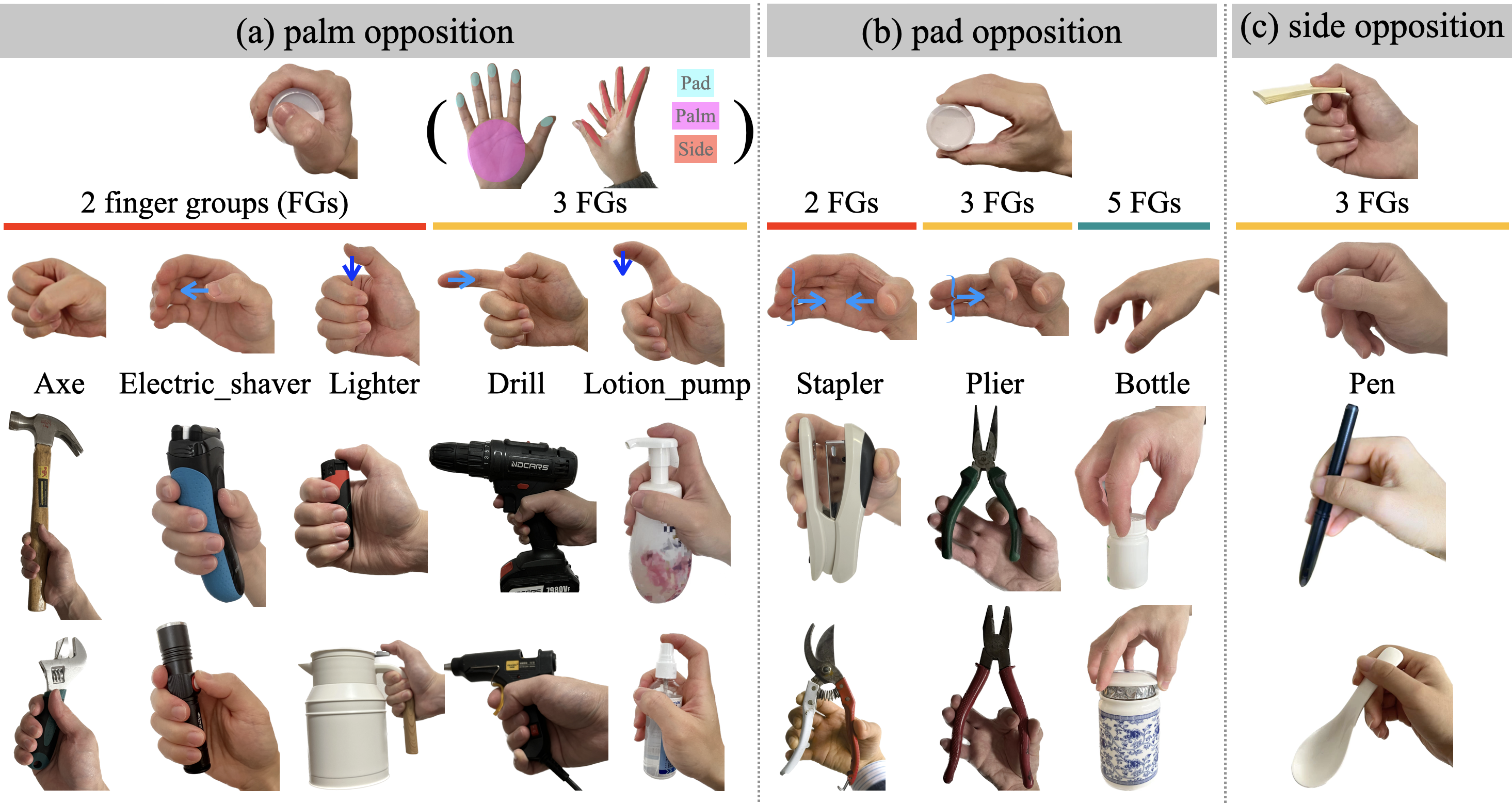}
\vspace{-0.3cm}
\caption{The nine functional grasp types named after representative objects, each shows two examples.}
\label{fig_pose}
\vspace{-0.5cm}
\end{figure}
\par \textit{Grasp type similarity.} Shape and layout similarities establish cross-category associations from an object-centric perspective. Additionally, associations across categories can also be constructed based on human-centric grasp types. The human hand has a high degree of freedom, but it is redundant for grasping daily objects. This has led to many studies on grasp type classification, such as the commonly used 33 grasp types~\cite{feix2015grasp}. However, these classifications mainly focus on static grasp rather than the functional grasp required for dynamic manipulation, such as the ``drill'' grasp type in Fig.~\ref{fig_pose} that cannot be classified into the 33 grasp types~\cite{feix2015grasp}. Considering that there is currently no taxonomy for functional grasps, we propose the following classification criteria. First, according to the type of opposition~\cite{iberall1997human}, functional grasps can be initially classified into three types: 1) palm opposition, wherein an object is grasped stably through the opposition between the palm and fingers (Fig.~\ref{fig_pose}(a)); 2) pad opposition, formed by two finger pads opposing each other (Fig.~\ref{fig_pose}(b)); 3) side opposition, where the finger side is involved, as in the pinching action shown in Fig.~\ref{fig_pose}(c) with the upper side of the index finger and the thumb pad. Second, each opposition type can be further subdivided according to finger movements. For instance, in the palm opposition examples in Fig.~\ref{fig_pose}(a), the five fingers gripping the `electric shaver' (column 2) can be divided into 2 groups: the thumb, which manipulates the switch, and the remaining four fingers that grip the handle. Similarly, the `drill' type (column 4) can be divided into three finger groups: the thumb, the index finger, and the remaining three fingers. Moreover, while the ``lotion pump'' type (column 5) has the same finger grouping as the ``drill'' type, the index finger moves in different directions, resulting in different groupings. 

\begin{figure}[!t]
\centering
\includegraphics[width=1\linewidth]{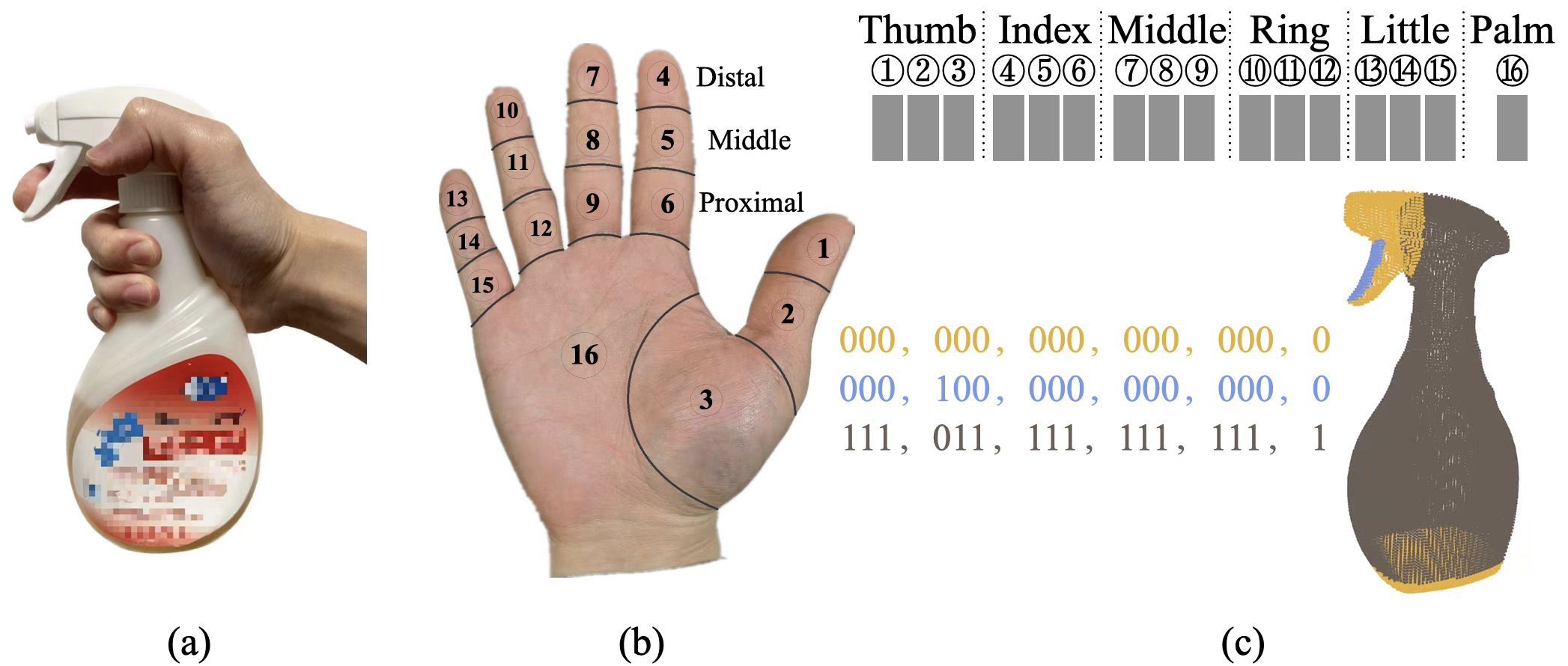}
\vspace{-0.3cm}
\caption{(a) An example of functional grasp. (b) The 16 parts of human hands. (c) A example of the touch code.}
\label{fig_touch}
\vspace{-0.3cm}
\end{figure}
\par To annotate the three types of similarities across different object categories, we first collect 1,768 objects across 51 categories from three sources: 1) self-reconstructed through 3D scanning, 2) self-collected from online vendors, and 3) existing datasets, including ShapeNet~\cite{chang2015shapenet}, YCB~\cite{calli2015benchmarking}, BigBIRD~\cite{singh2014bigbird}, KIT~\cite{kasper2012kit}, and Grasp Database~\cite{kappler2015leveraging}. As shown in Fig.~\ref{fig_teaser}(c), these object categories are organized into subgroups based on the hierarchical structure of WordNet~\cite{miller1995wordnet}, ensuring that each object within a category shares the same manipulation. For example, pliers are divided into `self-opening pliers' and `regular pliers' based on their different manipulation. We then invite 15 participants to manipulate these objects and vote on whether specific similarities exist across different object categories. Through this process, we grouped the 51 object categories into 12 groups for shape similarity, 17 groups for layout similarity, and 9 groups for grasp type similarity.

\subsection{Grasp Representation for Transfer}\label{sec:3.2}

\par With the three types of similarity mentioned above, this letter aims to transfer grasp on known objects to different categories of objects. Here, we choose to transfer touch code~\cite{zhu2023toward} because it is a semantic-based representation to specify which link of the dexterous hand should be used to contact the object part in a functional grasp, so it is not limited to human hands or specific mechanical dexterous hands. These characteristics are consistent with our desired generality requirements. An example of touch code is shown in Fig.~\ref{fig_touch}(c), where the sprayer bottle is divided into three parts, and each part has a 16-bit touch code to represent the correspondence between the object part and hand links (1 means that the hand link needs to touch this object part, and 0 means no touch). For example, the touch code of the blue trigger part is `000,100,000,000,000,0', indicating that this part is usually touched by the distal index finger. Since the existing dataset~\cite{zhu2023toward} contains insufficient object categories and quantities to support validation of cross-category touch code prediction, we expand the dataset by increasing the number of object categories from 18 to 51 and the number of objects from 129 to 1768. This expanded dataset can not only serve as the basis for functional grasp synthesis, but also be used as a benchmark to validate the performance of cross-category touch code prediction.
\vspace{-0.2cm}

\subsection{Architecture for Touch Code Transfer}\label{sec:3.3}
\begin{figure*}[!t]
\centering
\includegraphics[width=1\linewidth]{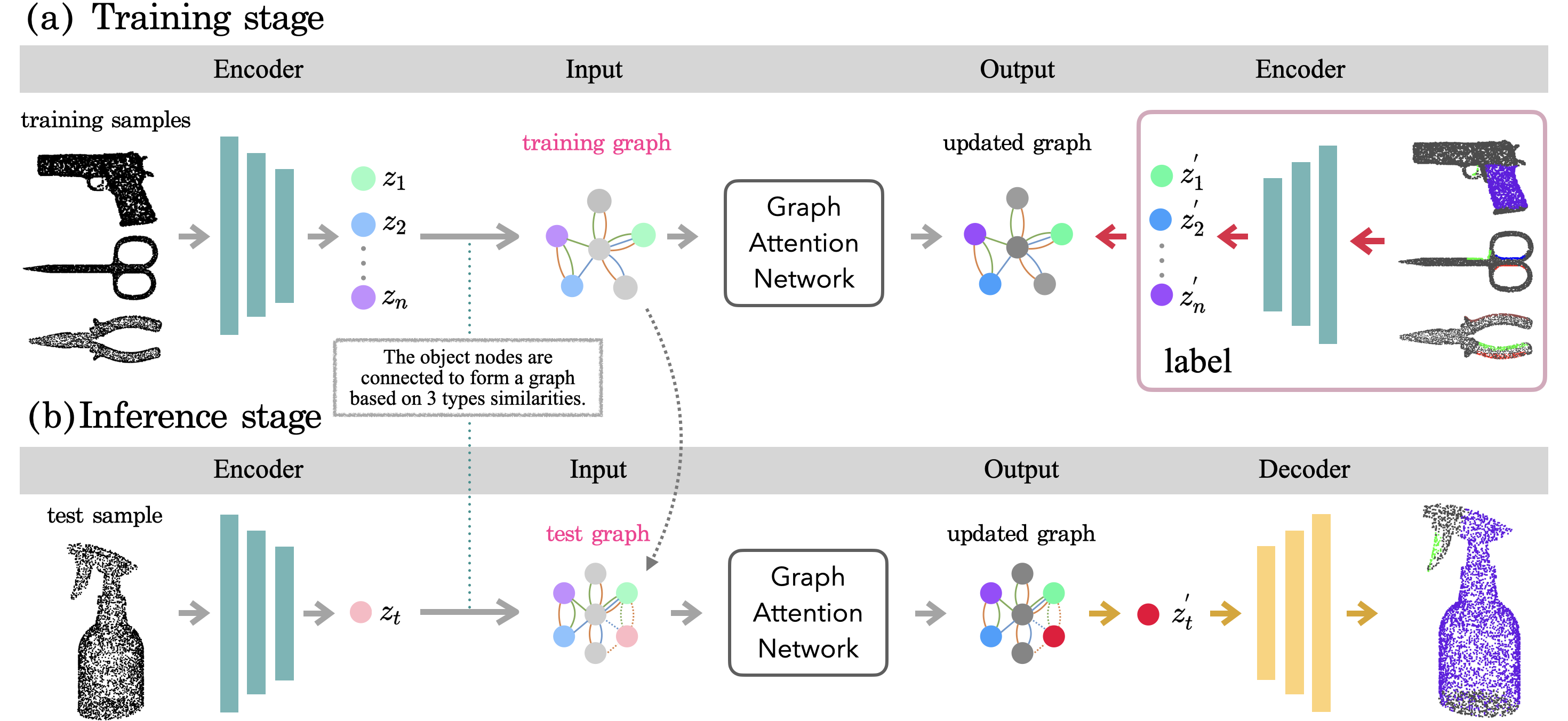}
\vspace{-0.2cm}
\caption{(a) Training stage of our prediction framework. (b) Inference stage for new objects.}
\label{fig_main}
\vspace{-0.2cm}
\end{figure*}
\begin{figure}[!t]
\centering
\includegraphics[width=1\linewidth]{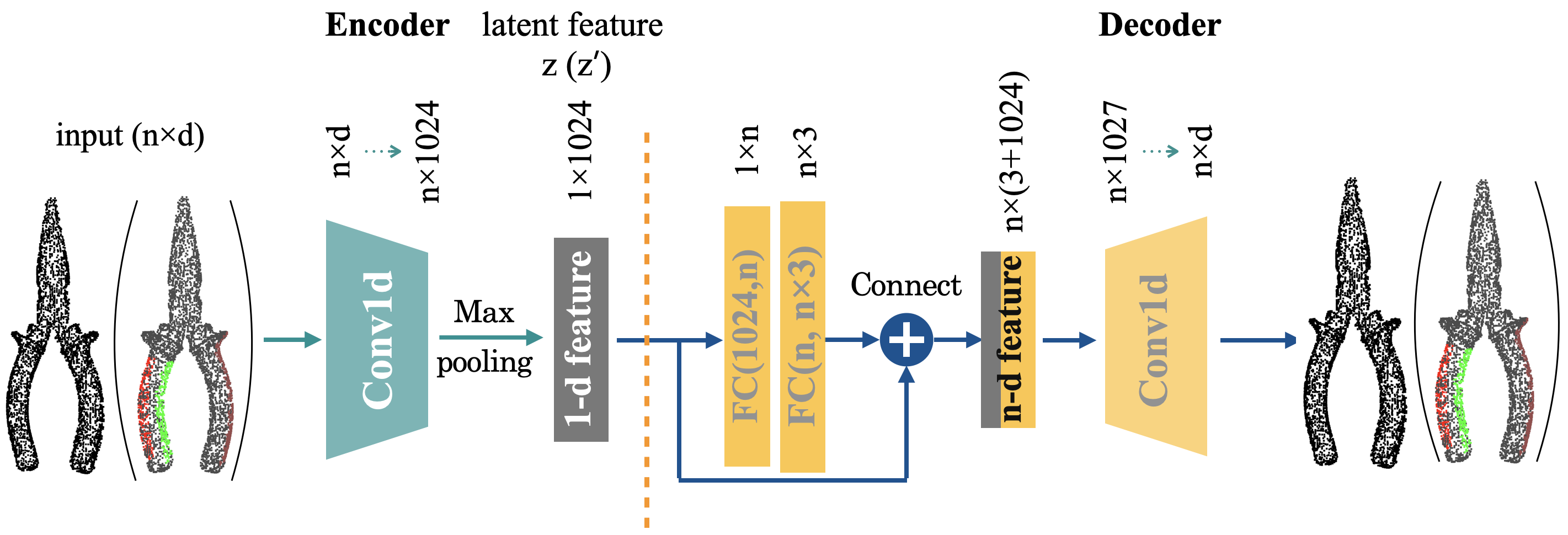}
\vspace{-0.7cm}
\caption{The auto-encoder network is designed to obtain vector-form features of the object point cloud or the object point cloud annotated with touch code. `n' is the number of object's surface points, and `d' is the feature dimension of each point. When the input is only the object's point cloud, `d' is 3, representing the coordinates of each point. When the input is point cloud annotated with touch code, `d' is 19, which includes the coordinates plus the 16-bit touch code.}
\label{fig_encode}
\vspace{-0.5cm}
\end{figure}

\par Since we have chosen touch code as the grasp representation in Sec.~\ref{sec:3.2}, this section aims to leverage three types of object similarities to transfer the touch code from known objects to new ones. Fig.~\ref{fig_main} illustrates the cross-category touch code transfer framework proposed in this section, which is divided into training and inference stages. In the training stage, the features of training objects are first extracted using an encoder, and then they are interconnected based on the three types of object similarities to form a training graph. A Graph Attention Network (GAT)~\cite{velivckovic2017graph} is then trained to predict the touch codes for all objects in the input training graph. During this phase, the GAT can see various training samples and learn how to infer touch codes based on similarity relationships. In the inference stage, the feature of the test sample is connected to the training graph according to the three types of similarities and then fed into the GAT along with the training samples. At this point, the trained GAT can leverage object similarities to transfer the touch code information learned from the training objects to the test sample. The following section will provide a detailed explanation of the core concepts and processes involved in these two stages.

\par When people grasp a new object, they subconsciously use the similarity between the new object and known objects to infer how to grasp the new object based on prior grasping experience with known objects. To mimic this human reasoning process, we have summarized three types of similarities to bridge cross-category objects in Sec.~\ref{sec:3.1}. However, we still lack a decision-maker capable of transferring touch codes across different categories. To address this, we employ the GAT as the decision-maker, as it can be trained to learn how to use the features of other related samples to update each sample's feature. Considering that the GAT requires working with vector-form features, we employ an auto-encoder network (AEnet) to obtain one-dimensional features of objects. The advantage of the AEnet is that the encoded feature can reconstruct the complete input of the object, allowing the GAT to fully observe the object based only on its feature, including global shape and local layout. Here, we use the AEnet to extract the following two types of one-dimensional features for training GAT. As shown in Fig.~\ref{fig_encode}, when the input to AEnet is only the object’s point cloud, the encoder outputs the object feature $z$ without touch information; when the input is the object point cloud with each point concatenated with the 16-bit touch code, the encoded feature is object feature $z^{'}$ with touch information.
\par Fig.~\ref{fig_main}(a) shows the training stage of our touch code transfer framework. To train GAT, we first organize the data into a graph structure. Here, we encode the point cloud of objects into features $\left \{ z_{i}  \right \} _{1}^{n} $, which serve as the initial features of the graph nodes. These nodes are interconnected based on the object similarities we proposed, with each edge corresponding to one of shape, layout and grasp type similarities. In this way, we construct the training graph shown in Fig.~\ref{fig_main}(a). Next we train the GAT to transfer hand-object touch information to the relevant nodes according to the 3 similarities. As shown at the right side of Fig.~\ref{fig_main}(a), we use known object features $\left \{ z_{i}^{'}  \right \} _{1}^{n} $ that include touch code information to supervise the output of the GAT. Through this supervised training, the GAT can learns how to transfer touch features from related objects. This is the foundation for achieving cross-category touch code transfer, where the GAT can infer the touch code of a new object based on the features of related objects that the GAT has seen. 
\par Fig.~\ref{fig_main}(b) shows the inference stage, where the pink node $z_{t}$ represents the encoded test object feature. First, we need to connect this new object's feature into the known training graph. As mentioned in Sec.~\ref{sec:3.1}, we have divided 51 categories of objects into 12, 17, and 9 groups based on shape, layout, and grasp type similarities, respectively. Based on this, we train three point cloud classification networks to predict the group of the new object in each type of similarity, and the new object is connected to learned objects within the same group. Then, we utilize the capability of GAT of handling dynamic graphs, i.e., the number of nodes in the graph does not need to be fixed. Thus, the new graph formed by adding a test node to the training graph can also be processed by GAT, and the pre-trained GAT can then rely on the three types of similarities to collect touch information from related objects to infer the new object's touch feature $z_{t}^{'} $ (red dot). This new object can be unseen or even from completely new categories. Finally, the predicted touch feature $z_{t}^{'} $ of the new object can be reconstructed to the object point cloud with touch code by the decoder of the pre-trained AEnet. In general, we have build the object associations through the knowledge graph and flexibly use GAT to handle dynamic graphs based on the three types of similarities to complete the prediction of a cross-category object’s touch code.

\par Some details are as follows. Our algorithm runs on a 2080Ti GPU. To balance computational cost and prediction accuracy, we construct a total of 90 knowledge graphs to train the GAT, with each graph consisting of 360 nodes and each node corresponding to a randomly selected one of the 1096 training objects from 36 categories. In the inference stage, we connect the test object feature to a random one of 90 training knowledge graphs, thus obtaining a test graph with 361 nodes. At this stage, we test a total of 672 objects from 49 classes, including 15 unseen classes, to test cross-category prediction performance of our method (Sec.~\ref{sec:4.1}).

\vspace{-0.2cm}
\subsection{Grasp Synthesis based on Transferred Touch Code}\label{sec:3.4}
\begin{figure}[!t]
\centering
\includegraphics[width=0.8\linewidth]{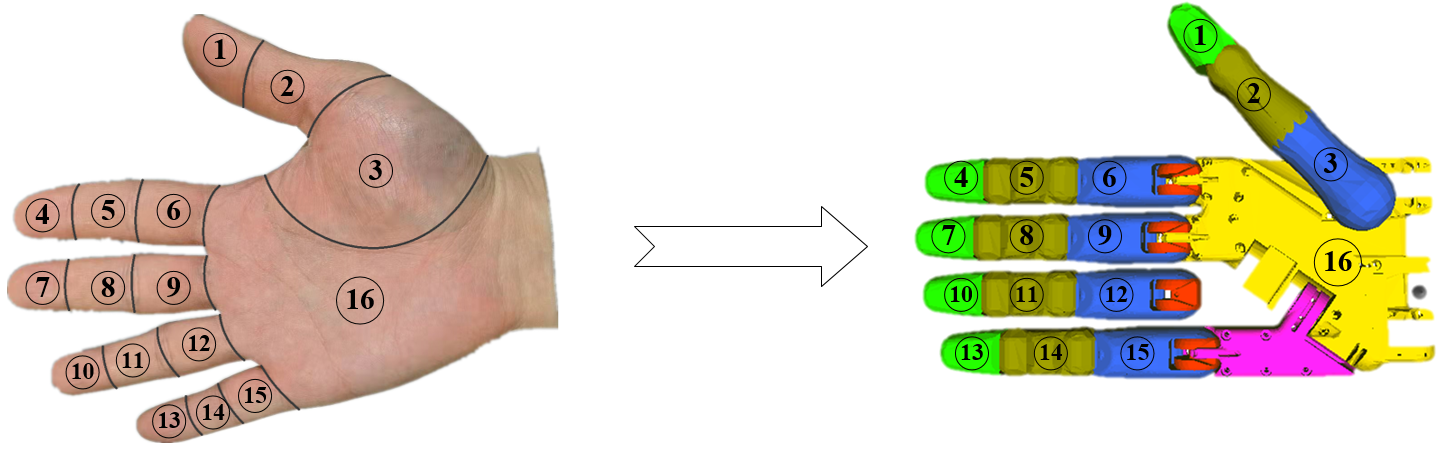}
\vspace{-0.3cm}
\caption{Correspondence between human hand and Shadowhand.}
\label{fig_map}
\vspace{-0.6cm}
\end{figure}

\begin{figure*}[!t]
\centering
\includegraphics[width=1\linewidth]{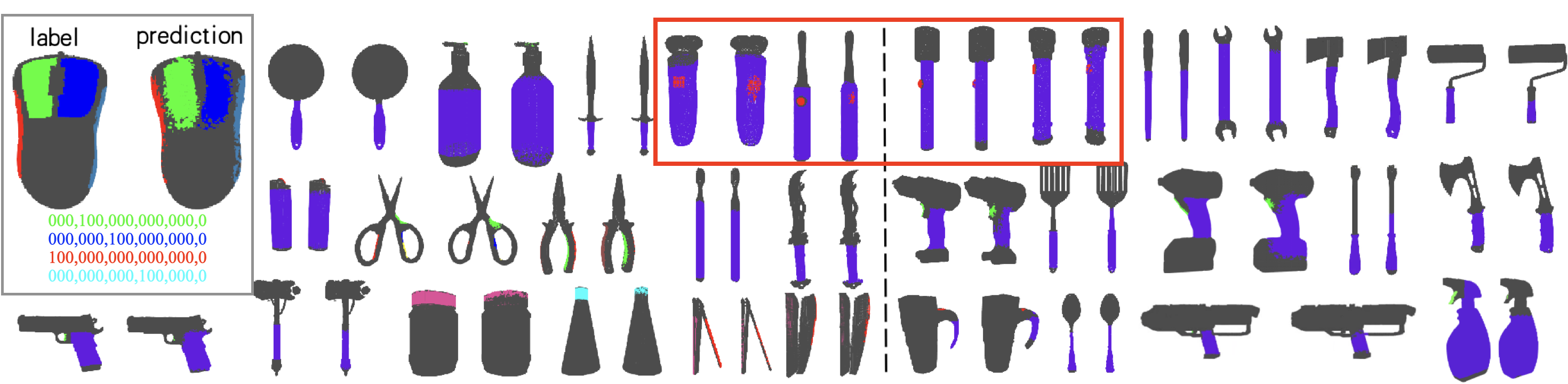}
\vspace{-0.7cm}
\caption{The prediction result of the touch code, with known categories on the left side of the dashed line and new categories on the right. Taking the mouse in the top-left corner as an example, each object displays the touch code label (left) and the predicted touch code (right), with different color regions representing distinct touch codes.}
\label{fig_obj}
\vspace{-0.3cm}
\end{figure*}

\begin{table*}[!h]
\centering
\caption{The mIoU results of 10 combinations, including the mIoU of seen and unseen categories of objects. S=Shape, L=Layout, G=Grasp type, S+L means the simultaneous use of shape and layout similarities. 
Avg=average mIoU.\label{tab:1}}
\scalebox{0.85}{
\begin{tabular}{cc|ccccc|ccccccc}
\hline
\multirow{2}{*}{Row} & \multirow{2}{*}{Relations} & \multicolumn{5}{c|}{Seen}                                                                    & \multicolumn{7}{c}{Unseen}                                                                                                         \\ \cline{3-14} 
                     &                            & hammer           & pan              & gun              & mug              & Avg              & flashlight       & scrather         & electric\_teapot & trigger\_sprayer & spatula          & paint\_roller    & Avg              \\ \hline
1                    & No                         & 66.66\%          & 90.16\%          & 40.19\%          & 82.52\%          & 59.28\%          & 51.03\%          & 60.28\%          & 40.37\%          & 12.25\%          & 68.55\%          & 40.12\%          & 44.34\%          \\
2                    & Shape                      & 71.55\%          & 96.88\%          & 51.81\%          & 88.42\%          & 68.96\%          & 61.33\%          & 65.85\%          & 52.30\%          & 35.15\%          & 78.64\%          & 58.07\%          & 58.29\%          \\
3                    & Layout                     & 70.26\%          & 94.37\%          & 47.07\%          & 87.96\%          & 66.11\%          & 60.18\%          & 64.25\%          & 46.02\%          & 39.34\%          & 75.20\%          & 56.89\%          & 55.28\%          \\
4                    & Grasp type                 & 69.89\%          & 95.09\%          & 51.76\%          & 86.01\%          & 67.92\%          & 61.46\%          & 65.39\%          & 48.21\%          & 37.83\%          & 76.66\%          & 62.08\%          & 57.92\%          \\
5                    & S+L                        & 72.29\%          & 97.43\%          & 56.74\%          & 92.72\%          & 70.60\%          & 63.83\%          & 66.42\%          & 50.82\%          & 42.18\%          & 81.19\%          & 62.14\%          & 62.23\%          \\
6                    & S+G                        & 72.57\%          & 97.01\%          & 53.76\%          & 92.28\%          & 72.69\%          & 63.50\%          & 68.58\%          & 59.17\%          & 40.70\%          & \textbf{81.84\%} & 71.05\%          & 62.47\%          \\
7                    & L+G                        & 73.58\%          & 97.25\%          & \textbf{57.56\%} & 92.16\%          & 72.83\%          & \textbf{67.02\%} & 69.18\%          & 58.22\%          & 41.91\%          & 80.46\%          & 72.87\%          & 63.55\%          \\
8                    & S+L+G                      & \textbf{75.75\%} & \textbf{97.52\%} & 57.47\%          & \textbf{92.92\%} & \textbf{73.49\%} & 66.96\%          & \textbf{72.33\%} & \textbf{60.36\%} & \textbf{45.36\%} & 81.19\%          & \textbf{76.83\%} & \textbf{64.02\%} \\ \hline
9                    & PointNet                   & 70.71\%          & 95.51\%          & 41.26\%          & 92.13\%          & 68.93\%          & 62.13\%          & 67.67\%          & 55.62\%          & 42.39\%          & 77.96\%          & 72.33\%          & 59.32\%          \\
10                   & PointNet++                 & 71.24\%          & 96.15\%          & 42.37\%          & 92.25\%          & 71.57\%          & 63.56\%          & 69.34\%          & 57.73\%          & 43.11\%          & 79.17\%          & 73.98\%          & 61.66\%          \\ \hline
\end{tabular}
}
\end{table*}
\par This section introduces how to synthesize the functional grasp $\left \langle J,T,R \right \rangle $ based on the transferred touch code $O_{c}$ and the functional grasp type $G$, where $J$ represents the joint angles of the dexterous hand, $R$ and $T$ respectively denote the rotation and translation of the hand relative to the object. To reduce the difficulty of optimizing the rotation variables in $SO(3)$, we first refer to ~\cite{murphf2021implicit} to provide a coarse rotation prior $R_{0}$ for the grasp synthesis, where the object point cloud feature and one candidate rotation prior (24 candidates in this letter, discretized along the main axes:$(x,y,z,-x,-y,-z)$) are input to the multilayer perceptron (MLP). The MLP then outputs the joint probability density, which can be used to select the optimal rotation prior. In this way, the grasp synthesis can be modeled as $\mathcal{M}:\left \{ O_{c},G,R_{0} \}\Rightarrow \{ \emph{J},\emph{R},\emph{T} \right \}$, and here we use the classical optimization method to synthesize the functional grasp. Specifically, we choose the Shadowhand as the tested hand because it is currently the most dexterous robotic hand with the highest degrees of freedom, and optimizing it requires considering more variables, making it difficult to synthesize functional grasp. 

\par To synthesize the functional grasp of an object, we first provide the optimization algorithm with a good initial pose of Shadowhand based on our functional grasp type ($G$). Specifically, we annotate nine pre-grasp poses for Shadowhand corresponding to nine functional grasp types. We then train the PointNet++ to predict the grasp type of the target object, thereby obtaining its pre-grasp pose. To make the touch code ($O_{c}$) applicable to Shadowhand, we select 16 links of Shadowhand corresponding to 16 parts of human hand as shown in Fig.~\ref{fig_map}. After completing the above steps, we refer to the loss functions from prior work~\cite{wu2023functional} and use three objective functions to optimize the functional grasp of the Shadowhand. The first is the attraction function, which minimizes the distance between the links of the dexterous hand and the related object area specified by the touch code (the area with touch code 1), such as the distance between the distal thumb and the flashlight switch. The second is the repulsion function, which serves to keep the links of the dexterous hand away from non-related object areas specified by the touch code (the area with touch code 0). The third is the hand regularization function, which consists of two parts. The first part keeps the joint angles of the dexterous hand within a reasonable range, and the second part prevents the fingers of the dexterous hand from colliding with each other. For detailed formulas, please refer to our prior work~\cite{wu2023functional}. 

\par Our grasp optimization is divided into two stages. The first stage only adjusts the translation and rotation of the hand, with an initial learning rate of 0.001 and a total of 100 rounds of optimization. The second stage mainly adjusts the joint angles of the dexterous hand, with an initial learning rate of 0.001, while the learning rates for translation and rotation are reduced to 0.0001, with a total of 200 optimization rounds. Thanks to the functional grasp types we proposed, this letter synthesizes higher-quality functional grasps using only classical optimization methods compared to existing network-based approaches~\cite{wu2023functional}.

\section{EXPERIMENT}

\label{sec:4}
\subsection{Result of Similarity-based Touch Code Transfer}\label{sec:4.1}
\par As described in Sec.~\ref{sec:3.3}, our framework relies on three types of inter-object similarity to transfer the touch code of the training objects to new objects. We test a total of 49 object categories, including 15 unseen categories used to evaluate the cross-category touch code prediction performance. Fig.~\ref{fig_obj} shows some of the test results, with seen categories on the left of the dashed line and the unseen categories on the right, where each object displays the touch code label (left) and prediction result (right) respectively. It can be observed that our framework not only predicts the touch codes of known categories effectively but also copes well with new categories. For example, the flashlight class highlighted in the red box is not involved in either the auto-encoder training or the GAT training, but associated with objects like electric shavers and electric toothbrushes through layout and grasp type similarities, enabling the flashlight to successfully achieve the touch code prediction of the head, switch, and body, without being misjudged as a hammer.

\par Moreover, Tab.~\ref{tab:1} presents the quantitative results for touch code prediction, using the evaluation metric of mean intersection over union (mIoU). `Row-1' represents the results of the baseline without using inter-object relationships, `Row-8' represents the results of our transfer framework when both training and testing use the correct shape, layout, and grasp type similarities. By comparing `Row-1' and `Row-8', it is evident that in both seen and unseen object categories, our method shows significant improvement over the baseline. We believe that when similarities are not used to associate objects, the network can only rely on its memory to handle some simple objects, such as 90.16\% mIoU of the pans in `Row-1'. However, it is incapable of handling more complex objects, such as guns, flashlights and trigger sprayers, whose mIoUs of `Row-1'(40.19\%, 51.03\%, 12.25\%) are significantly lower than `Row-8'(57.47\%, 66.96\%, 45.36\%). In addition, based on the well-trained model of `Row-8', 
we also evaluate the prediction effect when constructing test graphs using PointNet or PointNet++, where the shape, layout and grasp type similarities between the test object and known objects are predicted by PointNet or PointNet++. As shown in Tab.~\ref{tab:1}, the results of `Row-9' and `Row-10' are slightly inferior to `Row-8'. 

\par \textbf{Ablation Studies \uppercase\expandafter{\romannumeral1} .} We conduct ablation studies on three similarities (rows 2-8), and some interesting conclusions can be drawn from these results: 1) Based on Avg mIoU of both seen and unseen categories, the results can be roughly divided into three groups, where the multi-relations groups (rows 5-8) outperform the single-relation groups (rows 2-4), while the no-relation group (row 1) has a significant gap with the other two groups. Therefore, it can be concluded that good knowledge transfer requires sufficient associations between objects. 2) The performance of `L+G' (row 7) is better than `S+L+G' (row 8) in some objects, such as the flashlight. We speculate that for flashlights, the association based on layout and grasp type similarity with objects like electric shavers is more crucial. In this case, introducing associations based on shape similarity with objects like hammers might lead to interference. Nevertheless, it does not imply that shape similarity is not important, as adding shape similarity for most objects can improve accuracy. This inspires us to explore how to intelligently adjust associations between objects to improve performance next.
\begin{figure*}[!t]
\centering
\includegraphics[width=1\linewidth]{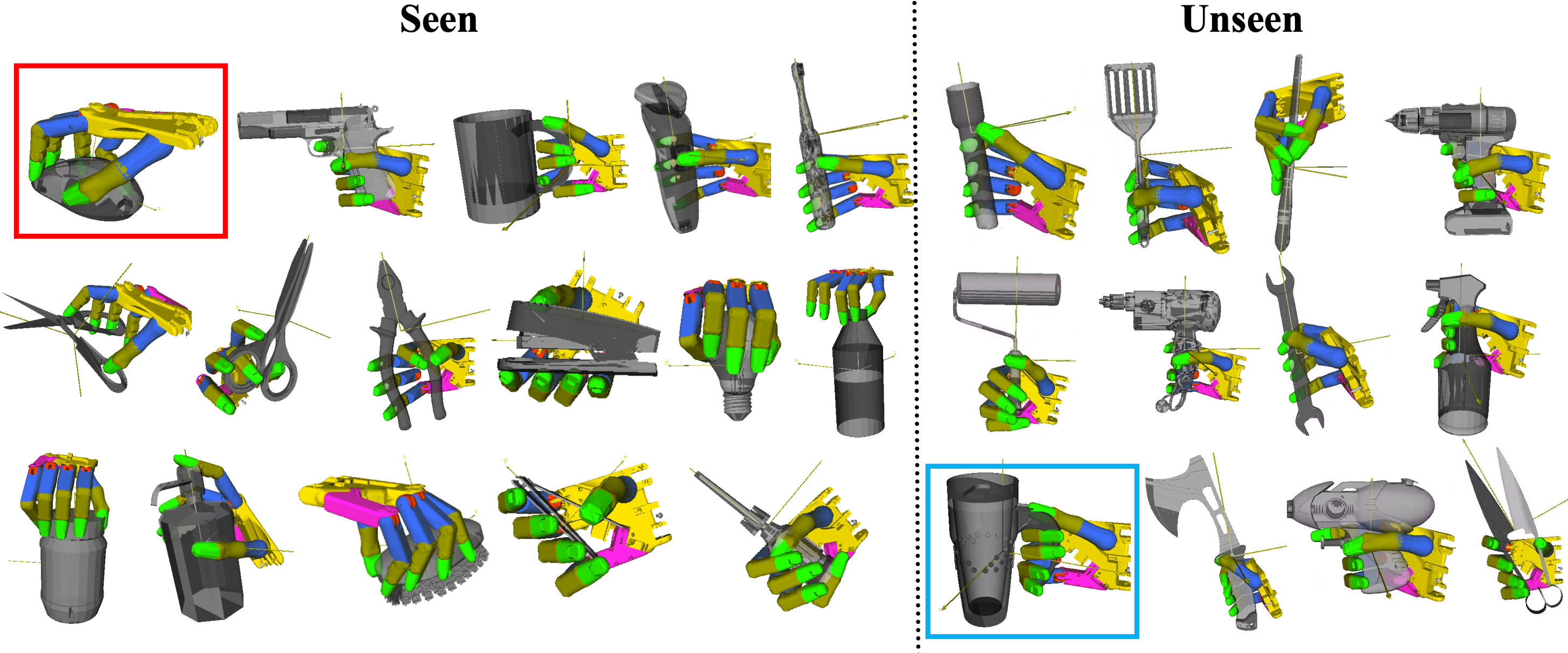}
\vspace{-0.7cm}
\caption{Functional grasp synthesis results based on predicted touch code.}
\label{fig_62}
\vspace{-0.2cm}
\end{figure*}

\begin{figure}[!t]
\centering
\includegraphics[width=1\linewidth]{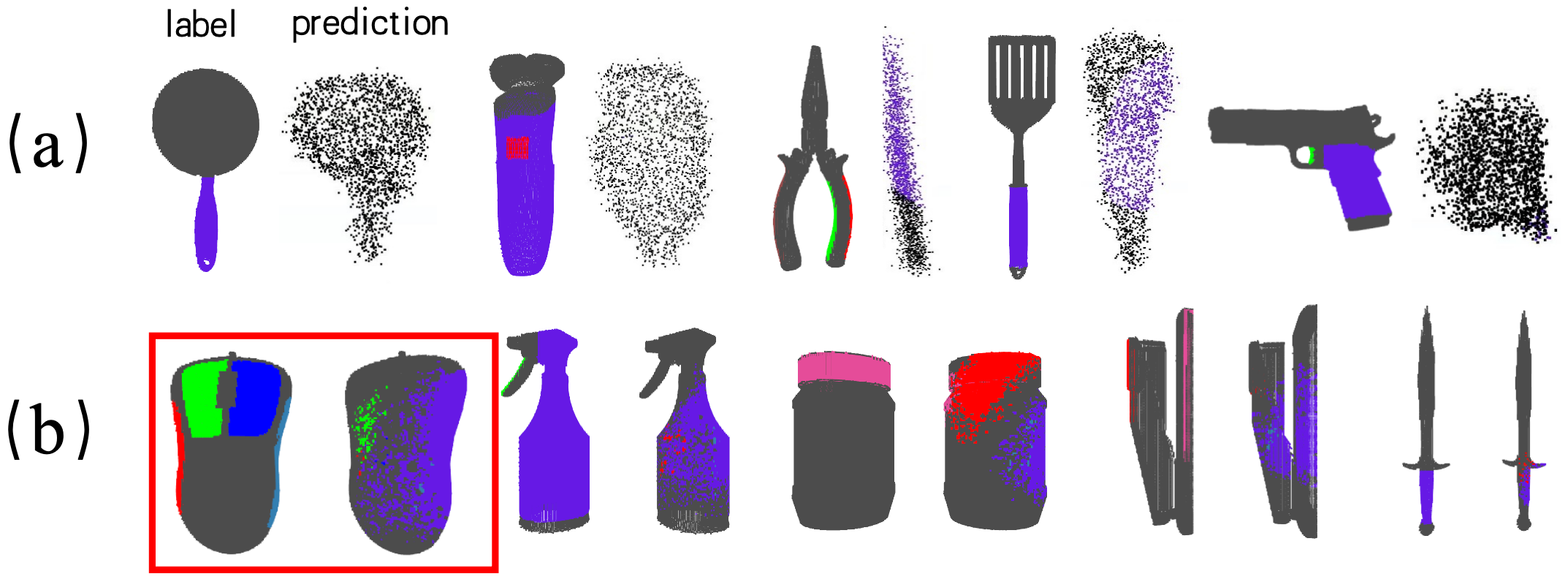}
\vspace{-0.6cm}
\caption{(a) The predicated results of network on the graph that containing only test samples. (b) The prediction results when randomly connecting test sample to learned samples.}
\label{fig_bad}
\vspace{-0.4cm}
\end{figure}

\par \textbf{Ablation Studies \uppercase\expandafter{\romannumeral2}.} In our inference stage, the test object is connected to the training graph. To verify whether the training graph (i.e., learned objects) are necessary for predicting touch code, we generate a 360-node test graph containing only test objects. Prediction results are shown in Fig.~\ref{fig_bad}(a), which reveal that the network is completely unable to handle these test samples, and even fail to retain the object structural information. Additionally, to further validate whether the similarities we proposed are effective in transferring the touch code, we randomly connect the test object to the previously learned objects in the training graph instead of using the correct associations, and the results are shown in Fig.~\ref{fig_bad}(b). Taking the first mouse as an example, its result is completely incorrect due to a mistaken association with objects like guns and drills, leading to incorrect touch code transfer. The above two experiments illustrate that our framework does not simply memorize the mapping between features, but actually achieves human-like reasoning, in which both reasonable inter-object associations and previously learned samples are crucial.

\vspace{-0.2cm}
\subsection{Functional Grasp Synthesis}\label{sec:4.2}
\par As mentioned in Sec.~\ref{sec:3.4}, we employ an optimization method based on predicted touch code and functional grasp type to synthesise functional grasp. It can be observed from the results in Fig.~\ref{fig_62} that objects from both seen and unseen categories can achieve functional grasps effectively. For example, within the red box, the dexterous hand's index and middle fingertip accurately land on the left and right buttons of the mouse, while the thumb, ring finger, and little finger grasp the sides of the mouse. Another example is the electric teapot in the blue box, where the thumb presses the switch above the handle, and the remaining fingers hold the handle. These successful functional grasp results stem from accurate touch code predictions and the effective pre-grasp poses provided by our functional grasp types. Moreover, we use Isaac Gym~\cite{makoviychuk2021isaac} to calculate grasp success rate. A successful grasp must lift the object against gravity for 5s and then withstand an additional 5s of continuously and randomly changing force disturbances. As indicated in the Tab.~\ref{tab:2}, our method demonstrate a good success rate on both seen and unseen objects. For instance, in the case of cross-category grasp synthesis, the success rate for flashlights reached 83.3\%. 
\begin{table}[!t]
\centering
\caption{Grasp success rates of our functional grasp.\label{tab:2}}
\vspace{-0.2cm}
\renewcommand{\arraystretch}{1.4}
\scalebox{0.65}{
\begin{tabular}{c|cccc|cccc}
\hline
\multirow{2}{*}{\begin{tabular}[c]{@{}c@{}}Success \\ Rate(\%)\end{tabular}} & \multicolumn{4}{c|}{Seen}     & \multicolumn{4}{c}{Unseen}                      \\ \cline{2-9} 
                                                                             & bottle & knife & mouse & gun  & drill & electric\_teapot & sprayer & flashlight \\ \hline
Ours                                                                         & 67.9   & 81.8  & 70.0  & 63.2 & 85.7  & 66.7             & 70.6    & 83.3       \\ \hline
\end{tabular}
}
\vspace{-0.2cm}
\end{table}

\begin{table}[!t]
\centering
\caption{Comparison with Existing Works. \label{tab:rbo}}
\vspace{-0.2cm}
\scalebox{0.9}{
\begin{tabular}{cccccc}
\hline
\multirow{2}{*}{Method} & \multicolumn{5}{c}{Grasp Success Rate}                  \\ \cline{2-6} 
                        & Drill   & Spray Bottle & Flashlight & Bottle  & Knife   \\ \hline
~\cite{rodriguez2018transferring}                 & 58.3\% & 57.1\%      & -          & -       & -       \\
~\cite{wu2023functional}                 & -       & -            & 83.8\%     & 71.7\%  & 90.5\%  \\
Ours                    & 85.7\%  & 70.6\%      & 83.3\%     & 67.9\% & 81.8\% \\ \hline
\end{tabular}}
\vspace{-0.4cm}
\end{table}

\subsection{Comparison with Existing Works}\label{sec:4.3}
\par To the best of our knowledge, the most relevant works to ours are~\cite{wu2023functional} and~\cite{rodriguez2018transferring}, both of which aim to achieve functional grasp transfer within a single category. Specifically,~\cite{rodriguez2018transferring} transfers grasp poses from one sample to other samples within the same category based on shape similarity, achieving grasp success rates of 58.3\% and 57.1\% on drills and spray bottles, respectively.~\cite{wu2023functional} transfers touch codes to other similar samples by densely matching surface points of objects within the same category, achieving success rates of 83.8\% on flashlights, 71.7\% on bottles, and 90.5\% on knives. Both methods require separate network to be trained for each object category. In contrast, our method is not limited to within-category transfer but instead uses a unified network model to achieve functional grasp transfer across different categories. In terms of grasp success rates, as shown in Tab.~\ref{tab:rbo}, our cross-category transfer method performs on par with in-class transfer methods, with some categories even surpassing them. For instance, the results for the drill and sprayer categories exceed those results in~\cite{rodriguez2018transferring}. In terms of inference time,~\cite{rodriguez2018transferring} requires an average of 7s, and~\cite{wu2023functional} takes 1.5s. In contrast, our method only needs 0.81s, with 0.05s for data processing, 0.16s for touch code transfer, and an average of 0.65s for grasp generation, demonstrating superior real-time performance.

\section{CONCLUSIONS}\label{sec:5}
\par To transfer the available functional grasps of objects to the other categories of objects, this letter analyzes the behavior of human grasping across categories and proposes three similarity relationships related to the grasp transfer. We propose the knowledge graph-based touch code transfer and optimization-based grasping synthesis framework, achieving cross-category functional grasp synthesis. It can be said that this letter creatively provides a new insights and new benchmark for cross-category functional grasp transfer. However, for the convenience of reader comprehension and method implementation, this letter employs relatively basic models. Therefore, there is still considerable room for improvement in performance. In the future, we will improve the network to enhance its performance and design a more flexible structure to further achieve grasping across tasks. By dynamically adjusting the links between nodes, we aim to enable objects to have multiple uses for different tasks.

\bibliographystyle{IEEEtran}
\bibliography{egbib}

\end{document}